\documentclass{IEEEmce}

\usepackage[colorlinks,urlcolor=blue,linkcolor=blue,citecolor=blue]{hyperref}

\usepackage{hyperref}
\hypersetup{ 
 colorlinks=true, 
 linkcolor=red, 
 filecolor=blue, 
 citecolor = blue, 
 urlcolor=blue, 
 } 
 
\usepackage{upmath}

\jname{Longevity, AI, and Cognitive Research Hackathon}

\begin{document}

\sptitle{Symbiosis project} 


\title{Designing an adaptive room for captivating the collective consciousness from internal states.}


\author{Sofía Vaca-Narvaja}
\affil{Brown University}

\author{Adán Flores-Ramírez, Ángel Alarcón-López}


\affil{Tecnológico de Monterrey / MIT }

\author{Daniela Leo-Orozco}
\affil{Tecnológico de Monterrey / Harvard }


\markboth{Symbiosis project}{Article Title}


\begin{abstract} Beyond conventional productivity metrics, human interaction and collaboration dynamics merit careful consideration in our increasingly digital workspace. This research proposes a conjectural neuro-adaptive room that enhances group interactions by adjusting the physical environment to desired internal states. Drawing inspiration from previous work on collective consciousness, the system leverages computer vision and machine learning models to analyze physiological and behavioral cues, such as facial expressions and speech analysis, to infer the overall internal state of occupants. Environmental conditions of the room, such as visual projections, lighting and sound, are actively adjusted to create an optimal setting for inducing the desired state, including focus or collaboration. Our goal is to create a dynamic and responsive environment to support group needs, fostering a sense of collective consciousness and improving workplace well-being. \textit{Keywords: Human Computer Interaction (HCI), Computer Vision (CV), Ubiquitous Technology, Human Connection}\end{abstract}

\maketitle

\enlargethispage{10pt}


\chapterinitial Research has consistently demonstrated our fundamental need for interpersonal connection, exemplified by intentional architectural designs like those of Apple and Pixar campuses, where Steve Jobs deliberately created spaces to foster spontaneous encounters among employees. The distinction between virtual and in-person engagement extends beyond mere preference; It has been documented fundamental differences in how our brains process faces during digital interactions compared to physical presence \cite{1}. Further research supports the value of group interactions, showing that groups with matched capabilities consistently outperform individual decision-makers and demonstrate superior processing capacity \cite{15}. Furthermore, the impact on creative output is measurable: It was found that face-to-face collaborating pairs generated 15-20\% more ideas than their virtual counterparts on Zoom, suggesting that digital platforms may impose previously unrecognized constraints on creative processes \cite{16}.

Inspired by the work done on Mediated Atmospheres at MIT’s Media Lab, we seek to potentiate the possibilities of ubicomp rooms by leveraging digitally enhanced features. Dr. Nan Zhao proposed a multi-modal workstation to create an immersive atmosphere for individuals, simulating a chosen environment to improve productivity in open-plan offices \cite{1}. We thrive to achieve an adaptive collaborative environment using multiple machine learning models to combine behavioral and biological markers of individuals as well as interactions, process them and reflect the feedback in physical changes of the environment (light, sound, visual projections and temperature). The goal is to capture a collective consciousness and create a unique metaphysic person from a group of physical people interacting in the same space.


\section{PREVIOUS WORK}

\subsection{On adaptive rooms}

Adaptive room research aims to combine the versatility of virtual environments with the capabilities of ubiquitous computing. The objective is to simulate the behaviors of future smart systems and enhance the potential of ubicomp rooms through digitally augmented objects. Designing adaptive rooms effectively requires addressing three key considerations: recognizing the cognitive and physical workflows present within the space, tuning the room’s configuration to meet the social needs of users during interactions, and ensuring coherence across environmental changes to avoid disorienting experiences. These principles emphasize that adaptive rooms should serve as comfortable, intuitive habitats rather than chaotic spaces reminiscent of “Alice in Wonderland” scenarios \cite{2}. The design philosophy for adaptive rooms rests on three foundational principles: prioritizing adaptation to human needs over requiring humans to adjust to the technology; incorporating smart objects and tools that collaborate to create a futuristic, digitally enhanced ubiquitous computing environment; and treating the room, along with its agents, as a system of distributed cognition where problem-solving occurs by coordinating both internal and external cognitive resources \cite{2}.

\subsection{On Inner State Recognition (behavioral + biological)}

The concept of internal states, as defined by SPIS, encompasses physiological, emotional, and other private experiences that are only accessible to the individual experiencing them \cite{3}. These include physiological conditions such as hunger or muscle tension, emotional experiences like love or pride, and private mental states such as motivations, wishes, and memories. The defining characteristic of these states is their subjective nature, as they cannot be objectively assessed by external observers. For example, only the individual can gauge how hungry they are, how much affection they feel, or the vividness of a particular memory. The definition also assumes that individuals can misinterpret their internal states, suggesting that self-assessment can be fallible. When considering whether one feels hungry, for instance, there is an implicit recognition that the individual might mistakenly identify or overlook their actual state \cite{3}.

Previous research on cognitive processes has predominantly focused on the role of neural networks at the individual level, with limited attention given to collaborative contexts. A collaborative approach emphasizes the interactions between individuals and examines how these interactions influence behavior and physiological parameters. 

Studies  have explored individuals' responses within controlled environments designed to achieve specific emotional states. In these settings, various factors such as electroencephalography (EEG), heart rate (HR), and electrodermal activity are measured and correlated with emotional states \cite{4}.   External stimuli are then adjusted to promote the desired emotional state.

Besides the emotional state of a person, another factor to analyze is their cognitive processes and how they influence their behavior. The engagement levels of an individual while performing an activity are closely correlated to this factor, systems can be trained with different sets of data to determine this factor. This can be achieved by analyzing not only physiological values, but also non verbal human behavior \cite{4}. Facial expressions like mouth and head movements, as well as eye behavior can be classified using different models to analyze the response of a person to an interaction with another individual \cite{5}. This has proven that emotional synchrony tends to occur when people are in each other's peripersonal space rather than far away \cite{6}. 
EEG signals also achieve a fundamental role in describing synchrony patterns between two individuals. When engagement is presented between two people, EEG synchronization takes place. This is fomented also by factors like visual contact and speech patterns \cite{7}.

\section{PROPOSAL}
The adaptive room needs a set of information to input into the feedback system. The proposed input will consist on the following physiological and behavioral aspects:

\subsection{Heart rate}

During active learning and engagement, heart rate typically exhibits momentary increases, characterized by spikes that rapidly return to baseline levels once the activity concludes \cite{8}. Heart rate measurement can be conducted through computer vision techniques, which use Eulerian video magnification for detecting rapid color changes in the person’s face to estimate beats per minute (BPM) \cite{9}.  Another approach is the use of thermographic imaging to monitor temperature variations associated with cardiac activity.

\subsection{Facial expression}

Emotional facial expressions tend to be similar among individuals in close physical proximity, facilitating emotional synchrony, particularly among younger participants \cite{10}. By analyzing facial cues, it is possible to infer shared emotional states and assess the degree of engagement.

\subsection{Body temperature}

Variations in body temperature can serve as indicators of emotional arousal, irritation, or anger. An increase in body temperature can be detected using thermographic sensors, providing a non-invasive measure of the individual’s emotional state.

\subsection{Respiratory frequency}

Respiratory frequency is another critical input, as it tends to rise and become more pronounced during states of arousal. This parameter can be monitored using wearable devices or computer vision techniques that assess chest or abdominal movements.

\subsection{Speech Analysis}

Speech analysis will involve sentiment analysis to detect emotional tone and evaluate engagement. In addition, the volume and tonal qualities of speech will be assessed, as these can provide insights into the speaker’s emotional state and level of engagement. \\

In designing environments that optimize user response and induce desired mental states, several key outputs can be adjusted based on real-time data and predictions from neural networks. These include visual projections, lighting, temperature, and auditory elements.

\begin{figure} [h!]
    \centering
    \includegraphics[width=1\linewidth]{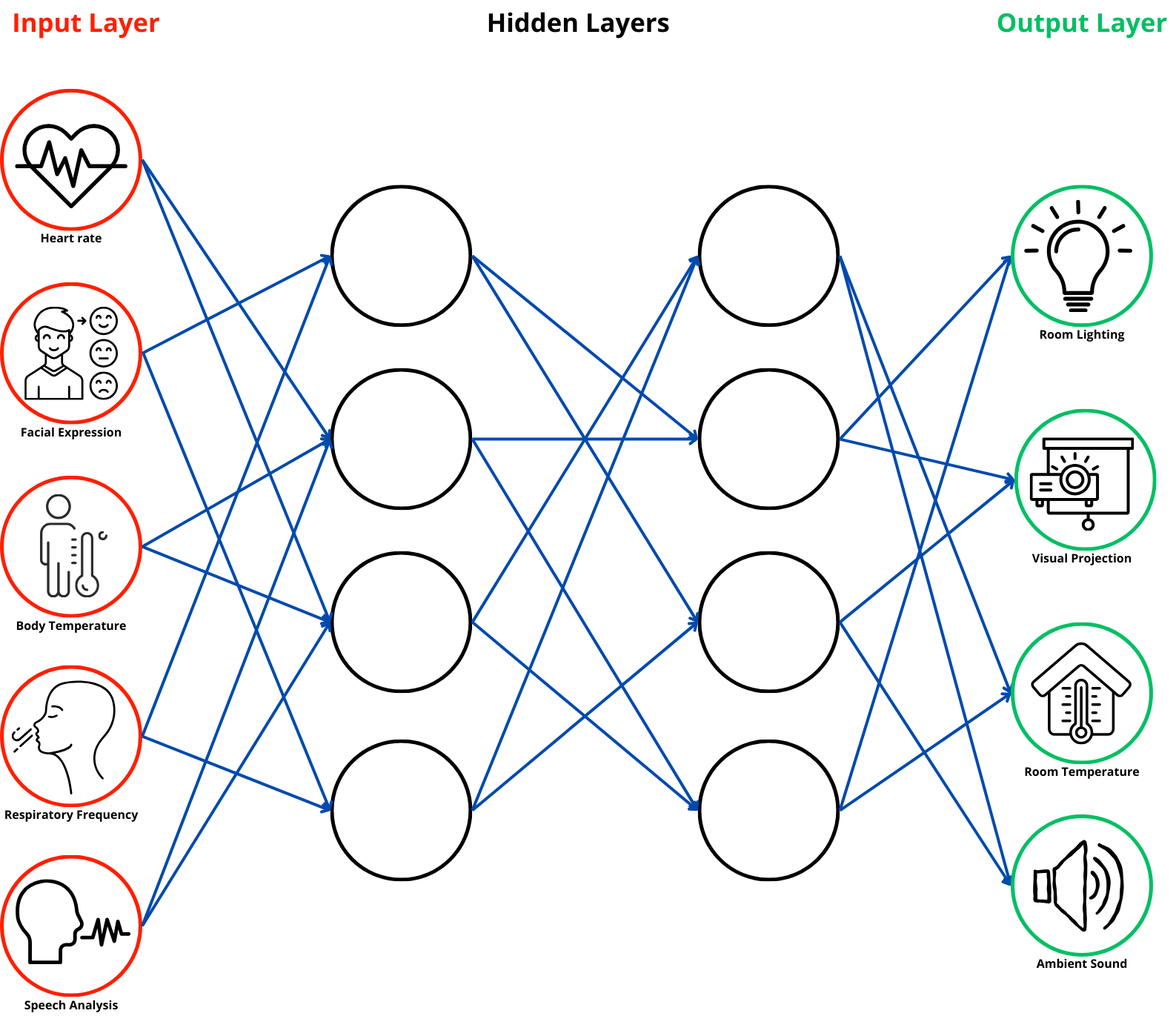}
    \caption{Neural network for output prediction, based on input on physiological and behavioral aspects of the human being}
    \label{fig:enter-label}
\end{figure}

\subsection{Visual projections}
Images and videos can be dynamically adjusted according to the neural network's predictions to create the optimal setup for specific work goals. This adaptive approach ensures that the visual environment aligns with the user's cognitive and emotional needs, enhancing focus and productivity \cite{11}.

\subsection{Lighting}

Lighting plays a crucial role in influencing emotional states. Heart rate variability is affected by different colored lighting conditions, indicating that specific wavelengths can alter emotional responses \cite{12}. By carefully selecting lighting conditions, it is possible to create environments that support desired mental states. 

\subsection{Temperature}

Studies have shown that mental workload increases when indoor thermal conditions deviate from a neutral state to either cooler or warmer settings \cite{13}. However, alertness tends to increase when the environment shifts from a neutral to a cooler condition. Therefore, maintaining an optimal thermal environment can enhance cognitive performance and comfort.

\subsection{Sound}

Auditory elements also significantly impact concentration and verbal reasoning. Both the type and intensity of sound can affect accuracy and efficiency. For instance, Liu et al. found that background noise types like intelligible speech negatively affect task performance compared to more favorable sounds like pure classical music \cite{14}. By modulating auditory inputs based on user needs, it is possible to create an acoustic environment that supports cognitive tasks and reduces distractions. 

By integrating these elements—visual projections, lighting, temperature, and sound—into a cohesive system informed by physiological data and neural network predictions, environments can be tailored to enhance user experience and performance effectively.

\section{ROOM SETUP}
\begin{figure} [h!]
    \centering
    \includegraphics[width=1\linewidth]{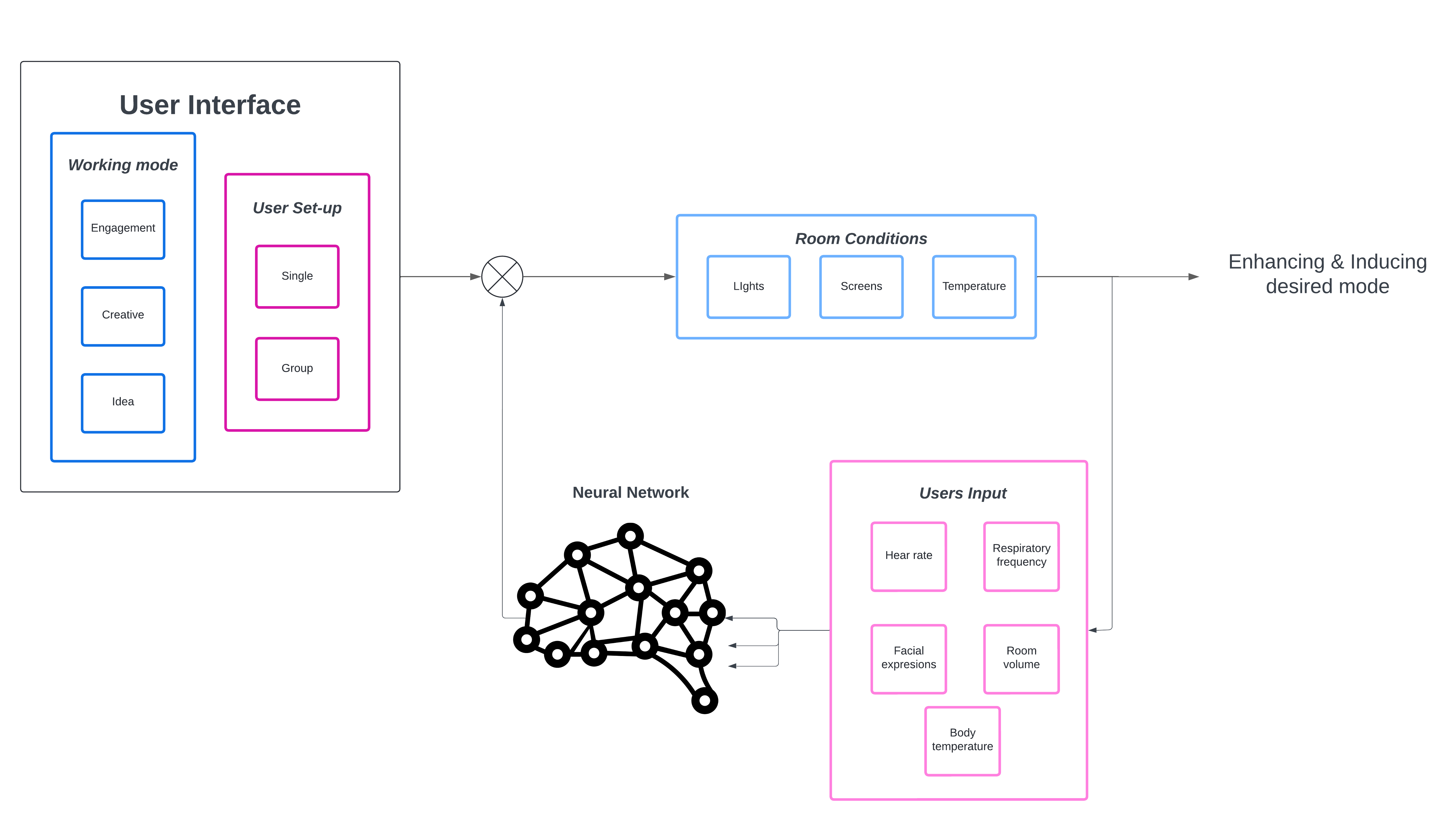}
    \caption{Feedback Loop system integrated with neural network for controlling output parameters}
    \label{fig:enter-label}
\end{figure}

The experimental environment is configured as an immersive space, integrating advanced display and sensory technologies to facilitate adaptive interactions. The room is equipped with OLED screens covering the walls and ceiling, providing high-resolution visual projections that can dynamically change according to the neural network's predictions for optimal work settings. This visual component is complemented by a 7.1 surround sound audio system, delivering an immersive auditory experience that enhances concentration and engagement. Furthermore, the environment incorporates a smart lighting system, which adjusts lighting conditions based on real-time data to influence emotional states positively. By maintaining optimal thermal conditions, the system supports cognitive performance and comfort. These components are interconnected through the Internet of Things (IoT), enabling seamless communication with a feedback loop system. This system processes inputs from strategically placed cameras to monitor user data non-invasively. Consequently, the room autonomously adapts to the occupants' needs, creating a responsive environment that enhances productivity and well-being.

\section{CONCLUSION}

This research explores the potential of adaptive environments to enhance group dynamics and well-being in shared workspaces. By integrating physiological and behavioral cues with real-time environmental adjustments, we aim to create a more personalized and productive space that fosters collaboration and a sense of collective consciousness.

\subsection{Future work}

Future work will focus on developing the adaptive room's physical design and refining AI models to create a responsive, inclusive environment. This space aims to enhance productivity and connection in offices, classrooms, and collaborative environments. By recognizing individual and group behaviors, the room could anticipate needs, suggest breaks, and adjust settings to foster creativity, inclusion, and collaboration, particularly supporting those with disabilities in communication and engagement. These advancements would enable a personalized, dynamic experience that adapts to and optimizes each user's needs.





\section{Annexes}
Github Room Generation Repo: \url{https://github.com/afr2903/neuroadaptive-meeting-room}


\newpage

\end{document}